\documentclass{article}
\usepackage{arxiv}
\usepackage[utf8]{inputenc} 
\usepackage[T1]{fontenc}    
\usepackage{hyperref}       
\usepackage{url}            
\usepackage{booktabs}       
\usepackage{amsfonts}       
\usepackage{nicefrac}       
\usepackage{microtype}      
\usepackage{lipsum}
\usepackage{pifont}
\usepackage{booktabs} 
\usepackage{graphicx}
\usepackage{listings}
\usepackage{color}
\usepackage{todonotes}
\usepackage{makecell}
\usepackage{float}
\usepackage{tabu, booktabs}
\usepackage[title]{appendix}
\usepackage{adjustbox}
\usepackage{array}
\newcolumntype{R}[2]{%
    >{\adjustbox{angle=#1,lap=\width-(#2)}\bgroup}%
    l%
    <{\egroup}%
}

\definecolor{dkgreen}{rgb}{0,0.6,0}
\definecolor{gray}{rgb}{0.5,0.5,0.5}
\definecolor{mauve}{rgb}{0.58,0,0.82}
\newcommand{\cmark}{\ding{51}}%
\newcommand{\xmark}{\ding{55}}%
\usepackage{xspace}

\newcommand{\sys}{RLzoo\xspace}
\usepackage{xcolor}
\definecolor{PrpRoyalBlue}{HTML}{4169E1}

\usepackage{tikz}
\newcommand*\myc[1]{%
\scalebox{0.78}{\begin{tikzpicture}[baseline=-3pt]
  \node[draw,circle,inner sep=0.5pt, fill=black] {\textcolor{white}{\textsf{\textbf{#1}}}};
\end{tikzpicture}}}

\title{Efficient Reinforcement Learning Development with RLzoo}
\author{
Zihan Ding\\
Imperial College London\\
\texttt{zd2418@ic.ac.uk} \\
\And
Tianyang Yu \\
Nanchang University\\
\texttt{tianyang2017@gmail.com} \\
 \And
Yanhua Huang \\
Xiaohongshu Technology\\
\texttt{iofficium@gmail.com} \\
 \And
Hongming Zhang \\
Peking University\\
\texttt{zhanghongming@pku.edu.cn} \\
 \And
Guo Li \\
Imperial College London\\
\texttt{lgarithm@gmail.com} \\
 \And  
Quancheng Guo \\
University of Edinburgh\\
\texttt{guoqch97@gmail.com} \\
 \And
Luo Mai \\
University of Edinburgh\\
\texttt{luo.mai@ed.ac.uk} \\
 \And
Hao Dong \\
Peking University \& Peng Cheng Laboratory\\
\texttt{hao.dong@pku.edu.cn} \\
}

\begin{document}
\maketitle

\begin{abstract}
Many researchers and developers are exploring for adopting Deep Reinforcement Learning (DRL) techniques in their applications. They however often find such an adoption challenging. Existing DRL libraries provide poor support for prototyping DRL agents (i.e., models), customising the agents, and comparing the performance of DRL agents. As a result, the developers often report low efficiency in developing DRL agents. In this paper, we introduce RLzoo, a new DRL library that aims to make the development of DRL agents efficient. RLzoo provides developers with (i) high-level yet flexible APIs for prototyping DRL agents, and further customising the agents for best performance, (ii) a model zoo where users can import a wide range of DRL agents and easily compare their performance, and (iii) an algorithm that can automatically construct DRL agents with custom components (which are critical to improve agent's performance in custom applications). Evaluation results show that RLzoo can effectively reduce the development cost of DRL agents, while achieving comparable performance with existing DRL libraries.
\end{abstract}


\section{Introduction}

In the last few years, we have seen many successful applications of Deep Reinforcement Learning (DRL) technologies, such as computer gaming~\cite{mnih2015human, lillicrap2015continuous}, robotic control~\cite{kober2013reinforcement, valassakis2020crossing}, self-driving cars~\cite{8957584}, language
modeling~\cite{park2018distort, furuta2019fully} and optimisation~\cite{li2017learning}.
To adopt DRL techniques, developers often need to construct DRL agents.
These agents interact with training environments, e.g., OpenAI Gym~\cite{brockman2016openai},  RLbench~\cite{james2020rlbench}, to collect training samples. Samples are sent to DRL algorithms which learn policies, e.g., on-policy algorithms~\cite{williams1992simple, schulman2015trust} and off-policy algorithms~\cite{mnih2015human}, 
that can maximise agent's performance, i.e., rewards,
in interacting with the environments.


Developing a DRL agent that can tackle a real-world application
is however challenging. There are several phases that are particularly time-consuming in developing a DRL agent: (i)~\emph{Prototyping phase}. A DRL agent contains various components (i.e., environments, DRL algorithms, and training drivers). To create a DRL agent, developers have to spend tremendous time in importing external training environments, modifying the environments to make them compatible with downstream DRL algorithms, and writing training drivers that can iteratively train the agents and distribute the training onto  heterogeneous processors~\cite{liang2018rllib}. 
(ii)~\emph{Customisation phase}. The default configuration (e.g., neural network architecture) of a DRL algorithm often exhibit sub-optimal performance~\cite{james2020rlbench, park2018distort}. Developers thus must customise the
DRL algorithm to improve its performance. 
(iii)~\emph{Algorithm comparison phase}. In tackling a training environment, there are often multiple DRL algorithms available~~\cite{williams1992simple, mnih2015human, schulman2015trust, heess2017emergence}. Developers usually need to implement all these algorithms and compare their performance. 
Even though several DRL libraries have become available recently,
developers still find it inefficient in using these libraries to construct DRL agents for custom applications.
On the one hand, tutorial-oriented DRL libraries, such as OpenAI Baselines ~\cite{baselines}, Stable Baselines and Coach~\cite{caspi_itai_2017_1134899}, provide command-line-based interfaces
 and they focus on reproducing classical benchmarks. They do not have low-level APIs which are necessary to control how a DRL agent is being trained, and how it is customised.
 On the other hand, research-oriented DRL libraries, such as  Tianshou~\cite{tianshou}, keras-rl~\cite{plappert2016kerasrl}, and Tensorforce~\cite{tensorforce}, provide
flexible APIs (e.g., defining the reward functions or policy networks), useful for defining custom DRL agents. They, however, fail to provide
expressive high-level APIs to help prototype DRL agents, and access commonly used DRL agents.

In this paper, we introduce \sys, a DRL library that can enable developers to efficiently prototype, train and evaluate DRL agents. The design of \sys makes several contributions:

\vspace{0.1cm}
\noindent
\textbf{(i)~High-level yet flexible APIs for declaring DRL agents.}
\sys contains high-level APIs for prototyping DRL agents. These APIs
contain \emph{expressive functions} for importing external training environments,
declaring DRL algorithms, and launching training drivers which can iteratively
train the policies and scale the training
to distributed nodes. 

Yet, \sys's APIs do not compromise flexibility.
They contain flexible functions which allow 
DRL agents to take custom agent components,  e.g., providing a
custom neural network for a DRL algorithm or providing a custom
communication topology for distributed DRL agents. 
By consolidating both the high-level and low-level APIs,
\sys is effective in facilitating both the prototyping phase and the customisation phase in developing DRL agents. 

\vspace{0.1cm}
\noindent
\textbf{(ii)~DRL model zoo.}
\sys provides a DRL model zoo to further facilitate
the algorithm comparison phase in developing DRL agents. 
The model zoo contains many useful pre-defined DRL environments and algorithms.
Generally, these algorithms can be classified into those for tutorial purposes (i.e., beginners)
and those for research purposes (i.e., professionals). In particular, 
\sys puts a focus on offering support for
 distributed DRL algorithms~\cite{heess2017emergence, espeholt2018impala} and  robot-learning-related environments, e.g., RLBench~\cite{james2020rlbench}. 

Further, the \sys model zoo contains an easy-to-use \emph{agent
training notebook}. \sys users can track the performance (e.g., reward) and configuration (e.g., hyper-parameters) of DRL agents, and evaluate different DRL agents in an intuitive manner.  

\vspace{0.1cm}
\noindent
\textbf{(iii)~Automatic algorithm for constructing DRL agents.}
\sys minimises developer's effort for integrating custom components 
into DRL agents, or re-configuring the agents for 
new scenarios. This is achieved by a novel 
algorithm that can automatically construct DRL agents with various custom components. Specifically, 
this algorithm has \emph{adaptors} for connecting the components (e.g., 
environments and DRL algorithms)
in DRL agents. The adaptors can automatically 
infer the input/output shapes of agent components.
As long as changes in these components are detected, the adaptors can automatically re-configure the DRL agent, which
avoids the need for developers to make manual modification 
as in existing DRL libraries.

\sys is implemented as a Python library based on TensorFlow~\cite{abadi2016tensorflow},
TensorLayer~\cite{dong2017tensorlayer} and KungFu~\cite{mai2020kungfu}.
It is open-sourced on Github\footnote{\href{https://github.com/tensorlayer/RLzoo}{https://github.com/tensorlayer/RLzoo}} 
in December, 2019. It has attracted numerous users from both education and industry. \sys has also been used for implementing demonstrations in a multi-lingual DRL textbook~\cite{dong2020deep}.

\section{RLzoo Design}

\label{sec:overview}

In this section, the design of RLzoo is introduced in detail. We start with an overview of the workflow, then describe its major APIs with a 
concrete code example.

\subsection{Overview}

\begin{figure}[H]   
\centering    
\includegraphics[scale=0.22]{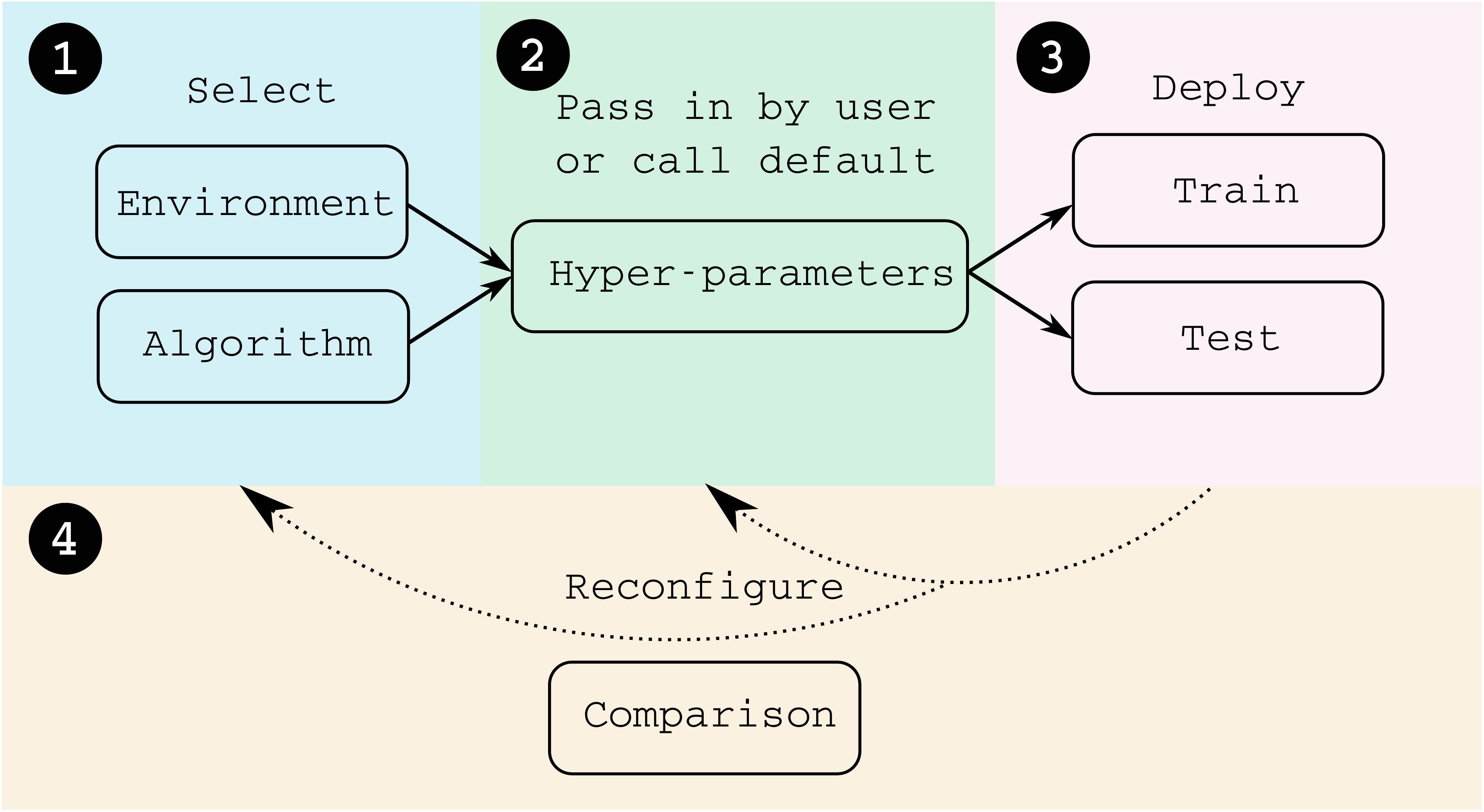}  
\caption{An overview of RLzoo usage with four key steps}
\label{fig:workflow}
\vspace{-0.5cm}
\end{figure}

The workflow of RLzoo is shown in Fig.~\ref{fig:workflow}. Four steps are necessary for training/testing: Step \myc{1}. Select the environment and the RL algorithm; Step \myc{2}. Pass in or call default hyper-parameters for the algorithm and learning process; Step \myc{3}. Start the training or testing. An additional step \myc{4} for re-configuration and iterative learning may be applied when comparisons among different algorithms and settings are required. 

\subsection{High-level yet Flexible APIs}
\begin{lstlisting}[float, language=Python, caption=Sample \sys program, numbers=left, floatplacement=t, captionpos=b, basicstyle=\ttfamily\small, label=alg:usage]
from rlzoo.common.env_wrappers import build_env
from rlzoo.common.utils import call_default_params
## Step 1: select and build the environment
env_type = 'classic_control'
env_name = 'Pendulum-v0'
env = build_env(env_name, env_type) # Build environment
## Step 2: choose the algorithm and get default hyper-parameters
from rlzoo.algorithms import TD3 # Choose algorithm
alg_params, learn_params = call_default_params(env, env_type, 'TD3') # Create configuration
## Step 3: create the RL agent and lauch learning process
agent = TD3(**alg_params) # Construct agent
agent.learn(env, 'train', **learn_params) # Launch training
\end{lstlisting}


The API design of \sys has the following goals: (i)~We want to 
enable users to use high-level expressive functions in declaring a DRL agent with custom
training environments, DRL algorithms and training drivers;
(ii)~We want to support users to flexibly customise their DRL agents by plugging different custom objects into DRL implementations. 

We introduce the \sys APIs using a sample program as shown in Listing~\ref{alg:usage}.
To declare a DRL agent,
\sys users first need to choose an environment (i,e., \texttt{Pendulum-v0}) (line~4$\sim$6). Based on the chosen environment, users further decide
 a DRL algorithm: \texttt{TD3} (line~8).
In order to train this DRL agent,
the users obtain its default construction parameters (line~11).
The algorithm parameters (\texttt{alg\_params}) are used
for constructing a DRL agent (line~8). 
This agent then launches its training process (line~12) given the environment and hyper-parameters (\texttt{learn\_params}).
As we can see in this program, a DRL agent can be declared with
3 abstracted steps (9 lines of code). 
A summary of API functions and descriptions are provided in Table~\ref{table:api}. The details of building the environments, importing the DRL algorithms, and constructing the DRL agents are hidden by the expressive API calls provided by \sys.
In the following, we will discuss the details of these API calls.

\begin{table*}[htbp]
\begin{tabular}{ |p{8.7cm}|p{7cm}|}
 \hline
 Function & Description \\
\hline
\makecell[l]{\textit{env} = \textbf{build\_env}(\textit{EnvName}, \textit{EnvType})} & Return the built environment instantiation with the name and type of it.\\
\hline

\makecell[l]{\textit{alg\_params}, \textit{learn\_params} = \textbf{call\_default\_params}(\textit{env}, \\ \textit{EnvType}, \textit{AlgName})} 
& Return two dictionaries of default hyper-parameters w.r.t. environments and algorithms. \\
\hline
\makecell[l]{\textit{agent} = eval(\textit{AlgName}+`(**\textit{alg\_params})') \\
\textit{agent}.\textbf{learn}(\textit{env}, \textit{mode}=`train', \textit{render}=False, **\textit{learn\_params}) 
}& 
\makecell[l]{Instantiate the class of DRL agent. \\ 
Launch training/testing process with the agent. \\
}\\
\hline
\end{tabular}
\caption{RLzoo API} 
\label{table:api}
\end{table*}

\vspace{0.1cm}
\noindent
\textbf{Building learning environment.} DRL environments
are often imported from external libraries, e.g., OpenAI Gym.
To hide the difference in the APIs of using these libraries,
\sys provides an abstracted 
function: \texttt{build\_env()} for importing
environments. This function takes the environment name \texttt{env\_name}
and its type \texttt{env\_type} (line~6 in Listing~\ref{alg:usage}). It automatically builds the environment by manipulating external libraries
and transparently operates this environment within the DRL agent.

\vspace{0.1cm}
\noindent
\textbf{Obtaining default agent configuration.}
Initialising a DRL agent needs massive parameters (e.g.,
the parameters for instructing the DRL algorithms
and those for controlling the training process).
Letting \sys users
decide all these parameters makes \sys difficult to adopt among
general users.
To address this, 
\sys provides default pre-tuned parameters for its DRL agents.
These parameters are obtained through the
\texttt{call\_default\_params()} function.
This function returns the pre-tuned algorithm parameters in the
dictionary \texttt{alg\_params}, and the learning hyper-parameters in the 
dictionary \texttt{learn\_params}.

\vspace{0.1cm}
\noindent
\textbf{Customising agents.}
\sys allows users to easily customise DRL agents.
This is achieved through providing an intuitive manner
for configuring the algorithm 
and learning hyper-parameter dictionaries. For example, to customise the neural
networks used within the DRL algorithm,
users can access the default neural networks
through the key: \texttt{`net\_list'}.
They can replace the default networks with custom neural networks. These neural networks
follow a shared \texttt{`Model'} interface. They can thus be seamlessly
integrated within the agent environment.
Following the same manner,
\sys users can customise other algorithm parameters, e.g., optimisers,
and learning hyper-parameters like learning rate and
batch size.

\vspace{0.1cm}
\noindent
\textbf{Constructing and manipulating agents.}
\sys makes the construction and manipulation of DRL agents
easy and efficient.
All DRL methods in \sys can be instantiated to be an \texttt{agent} (line 11 in Listing~\ref{alg:usage}).
This allows these agents to be manipulated consistently. All these
agents can share utility functions that
are pre-implemented within the base agent class, such as
the \texttt{learn()} function which launches
the training process or evaluates the performance of the agents.

\subsection{Automatic Agent Construction}
\label{sec:agent_construction}
Using \sys APIs, users
can implement a wide range of DRL agents.
This however introduces challenges.
These agents usually consist of 
 external environments and custom agent's modules, while those environments and modules may not 
be compatible
with the existing components in the \sys library
(e.g., some environments might produce image observations while the other produce vectors).

\begin{figure}[t]
\centering    
\includegraphics[scale=0.18]{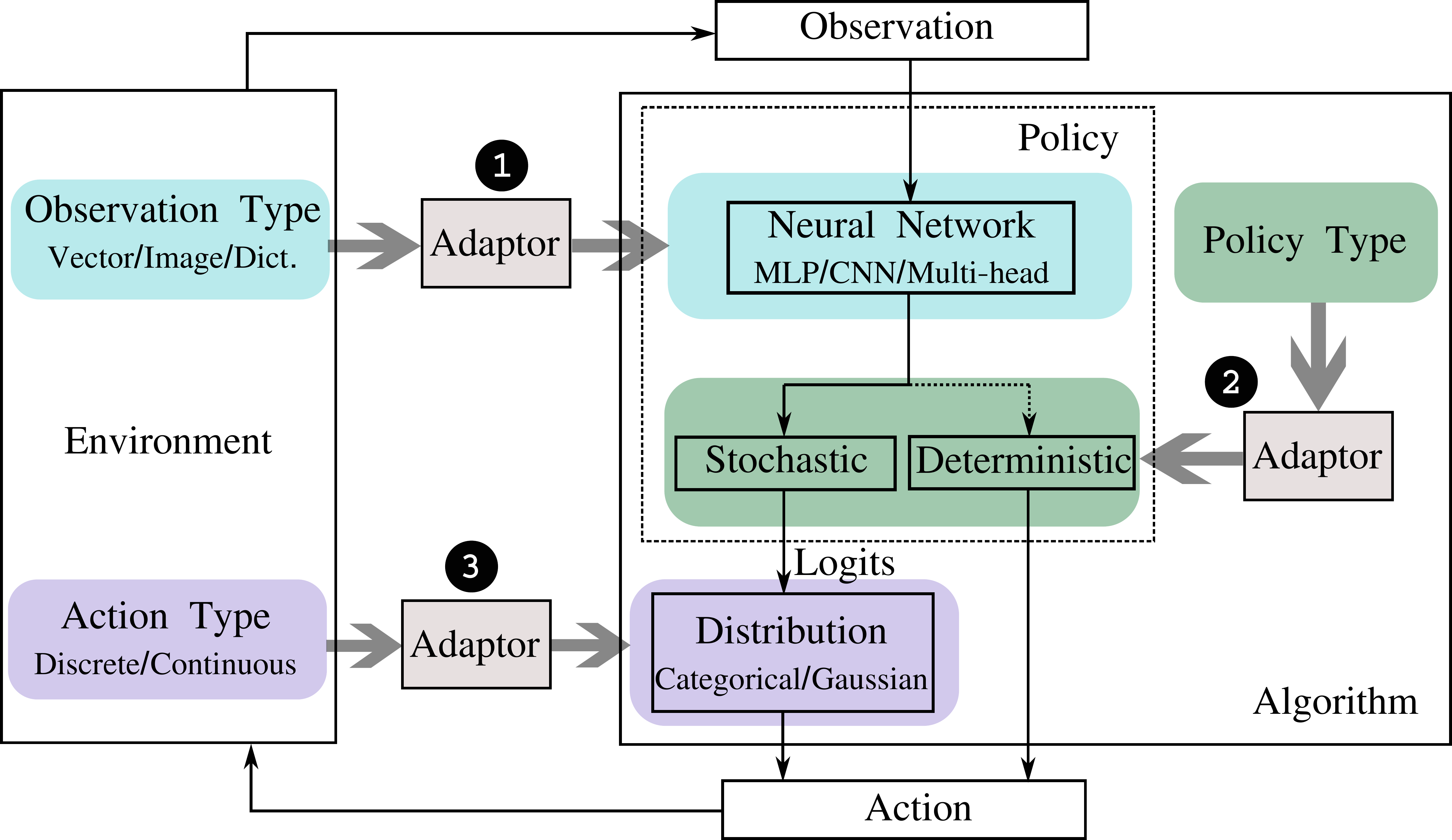}  
\caption{Automatic construction of a DRL agent.}
\label{fig:automodel}
\vspace{-0.5cm}
\end{figure}

Existing DRL libraries often rely on users to \emph{manually} resolve
this compatibility issue.  
Figure~\ref{fig:automodel} contains a typical architecture of a DRL agent (ignoring the three adaptors added by RLzoo).
In this agent, if a new environment is provided, 
users would have to manually update the neural
networks (e.g., MLP and CNN) to process the new observations.
This manual update needs to propagate through the agent, e.g., updating the policies, then the action types.
Relying on users to manually make all these updates
incurs high development costs.
It can also make the expertise of DRL become
a prerequisite for using DRL libraries.


\sys wants to automate the process of constructing
a DRL agent. Our key idea 
is to embed numerous \emph{adaptors} between agent components. 
These adaptors infer the type of the output from upstream
agent components and dynamically compute the input for downstream components.
The adaptors will be called in sequence and automatically re-configure the entire agent.

  Figure~\ref{fig:automodel} illustrates the automatic agent construction algorithm in \sys. First of all, this algorithm places
an \emph{observation adaptor}
between the observation from the environment and the neural network (see \myc{1}). This adaptor infers the type of observations and produces different types of neural networks,
e.g., a MLP for a vector, the CNN for an image,
and a multi-head architecture for a hybrid dictionary.
The \emph{policy adaptor} is placed in between
the algorithm selection and the policy output (see \myc{2}). Based on the stochastic/deterministic nature of the DRL algorithm, this adaptor produces corresponding policy outputs for each selected DRL algorithm.
The \emph{action adaptor} is placed in between the policy (stochastic only)
and the environment (see \myc{3}):
if the environment requires discrete action,
this final adaptor will produce a categorical distribution to represent the action; if the action needs to be continuous, the adaptor produces policy output as a diagonal Gaussian distribution.

\subsection{Model Zoo}

Making \sys useful for users
introduces unique challenges.
First, these users may not have the expertise of DRL to implement agents that are suitable for their applications.
Second, training the agents to reach high accuracy
often requires extra DRL knowledge in comprehending
training results, and tuning hyper-parameters. 

To address these challenges,
we want to make \sys a comprehensive DRL platform.
To avoid making DRL expertise
a prerequisite, \sys provides 
a large collection of pre-defined DRL environments
and algorithms which can be directly leveraged
by users. 
\sys makes the tuning of DRL agents easy for non-experts of DRL. This is achieved through
an interactive training notebook.
This notebook provides useful functionalities
for managing DRL agents and tuning their performance.

\vspace{0.1cm}
\noindent
\textbf{Pre-defined DRL algorithms and environments.}
Our choice for pre-defined
DRL environments and algorithms is driven
by the following observation:
many users like to 
first verify the benefits of DRL technologies 
by starting with simple DRL algorithms and environments.
They will gradually move to advanced DRL algorithms/environments once they realise the need for
improving the performance of DRL agents.

\sys supports both simple and advanced training environments.
Many simple environments have been integrated
within the library, including Atari, Box2d, Classic control, MuJoCo, Robotics in OpenAI Gym, and DeepMind Control Suite. These environments cover most of the classical DRL benchmarks we are aware of.
In addition, developers can access to more up-to-date environments such as those used for emerging realistic robot learning, e.g., RLBench~\cite{james2020rlbench}. These environments produce complex observations represented as compound dictionaries, and they can be used by practitioners to test DRL with robots.



\sys provides a large number of DRL algorithms.
Classical DRL algorithms, including the Deep Q-Network (DQN)~\cite{mnih2015human}
 and its variants~\cite{van2016deep, wang2015dueling, fortunato2017noisy, schaul2015prioritized} in the discrete action spaces,
 are pre-implemented in \sys.
 Many state-of-the-art DRL algorithms,
 which often achieve better agent performance,
 are pre-implemented as well.
 Examples include
hindsight experience replay (HER)~\cite{andrychowicz2017hindsight}, deep deterministic policy gradient (DDPG)~\cite{lillicrap2015continuous}, twin delayed deep deterministic policy gradient (TD3)~\cite{fujimoto2018addressing}, soft actor-critic (SAC)~\cite{haarnoja2018soft}, advantage actor-critic (A2C)~\cite{mnih2016asynchronous}, asynchronous advantage actor-critic (A3C)~\cite{mnih2016asynchronous}, 
proximal policy optimisation (PPO)~\cite{schulman2017proximal}, distributed proximal policy optimization (DPPO)~\cite{heess2017emergence}, trust region policy optimisation (TRPO)~\cite{schulman2015trust}.

\vspace{0.1cm}
\noindent
\textbf{Agent training notebook.} \sys users can exploit a high-level
agent training notebook to manage the configurations of DRL agents,
analyse their training performance metrics
This notebook is implemented based on the Jupyter Notebook. 
It tracks the configurations of being evaluated DRL agents
and stores the agents' traces for performance analysis.
Specifically, the notebook displays agent performance metrics, including agent configurations, learning status (e.g., training steps and instant rewards), and training results (e.g., averaged rewards over time and values of loss functions). 
Based on these metrics, developers can infer
the impacts of modification made towards the DRL agents,
thus facilitating the tuning of such agents for better performance. 

\subsection{Distributed Agent Training}
DRL agents usually need to accelerate computation using parallel heterogeneous processors. To achieve this, developers often rely on
the native \emph{multiprocess} library in Python. The usage of such a library, however, is generally limited within a single machine. To use distributed machines, developers must use
external libraries, such as Ray~\cite{moritz2018ray} and Acme~\cite{hoffman2020acme}. These libraries provide \emph{custom} Remote-Process-Communication (RPC) programming interfaces. Developers must largely modify existing DRL programs in order to adopt these RPC libraries.

In designing \sys, we want to minimise developer's efforts in modifying existing single-node programs when scaling out DRL agents. Our idea is to ensure \sys's distributed training APIs can follow the convention in the API design of the popular multiprocess library. For each key component in the multiprocess library (e.g., queues and pipes), \sys provides \emph{equivalent distributed implementations}. Hence developers can easily replace single-node communication components with those equivalent in \sys. Further, \sys extends the APIs of the multiprocess library. It provides novel collective communication APIs (e.g., all-reduce) to achieve complex communication patterns among DRL agents~\cite{mai2020kungfu}. 

\begin{figure}[t]
\centering    
\includegraphics[scale=0.49]{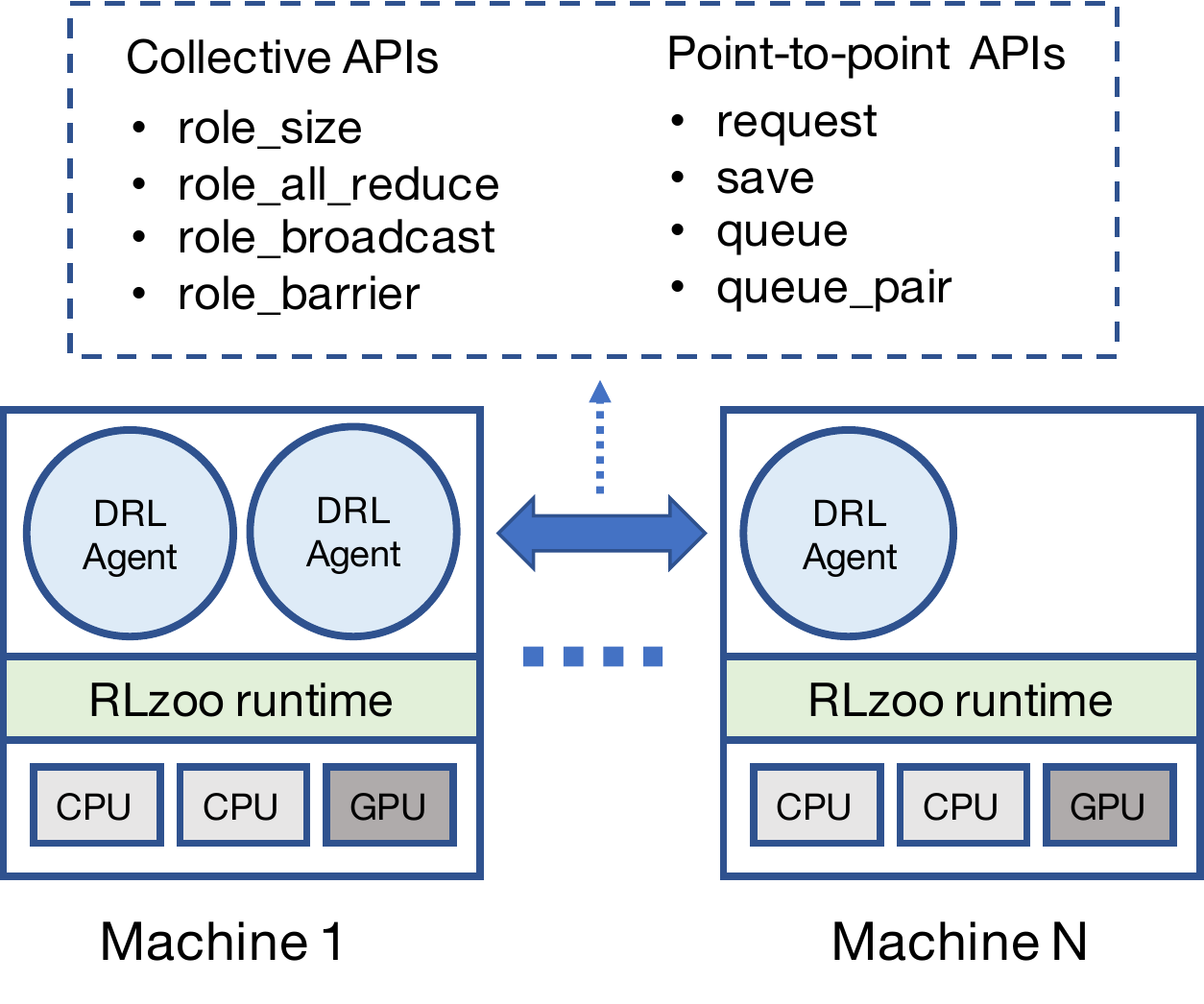}  
\caption{Distributed training architecture of \sys agents.}
\label{fig:distributed}
\end{figure}

Figure~\ref{fig:distributed} shows the distributed architecture of \sys DRL agents. 
In a cluster, \sys replicates \emph{RLzoo runtime} on each machine. This runtime
launches \sys DRL agents as Python processes, and assigns CPUs and GPUs to the agents. The DRL agents communicate data using expressive communication functions defined
in the \emph{collective APIs} and \emph{point-to-point APIs} (Figure~\ref{fig:distributed}):

\noindent
\textbf{Agent collective APIs}: \sys agents are often assigned with different roles (e.g., actors, learners and inference servers as in IMPALA~\cite{espeholt2018impala}). The agents in the same role can use (i) \emph{role\_all\_reduce} to synchronise the gradients among parallel
learners, (ii) \emph{role\_broadcast} to broadcast weights to parallel learners (which produce gradients to update DRL models), and (iii) \emph{role\_barrier} to coordinate the synchronous execution of parallel actors (which collect training trajectories from environments). 

\noindent
\textbf{Agent point-to-point APIs}: Developers can use (i) \emph{save} and \emph{request} to asynchronously push and pull weights among DRL agents, and (ii) \emph{queue} or \emph{queue\_pair} to exchange data among those agents, similar to the queues in the Python multiprocess library. 

\section{Evaluation}


In this section, we compare \sys with other
DRL libraries in terms of the supported algorithms, supported environments
and their API designs.
We choose the following popular libraries as baseline: OpenAI Baselines~\cite{baselines}, Tianshou~\cite{tianshou}, Coach~\cite{caspi_itai_2017_1134899}, ReAgent~\cite{gauci2018horizon}, garage~\cite{garage}, keras-rl~\cite{plappert2016kerasrl}, MushroomRL~\cite{deramo2020mushroomrl} and Tensorforce~\cite{tensorforce}.

\begin{table}[htbp]
\centering
\begin{tabular}{ |p{2.5cm}|p{1.9cm}|p{1.cm}p{1.cm}p{1.cm}p{1.cm}|p{1cm}|}
 \hline
 Library & \# Algo. & \# Env. & Image & Vector & Dict. & LoC \\
 \hline
RLzoo & 12 & 7  & \cmark & \cmark & \cmark & 4 \\
\hline
Baselines & 9 & 5 & \cmark & \cmark & \cmark & N/A\\
\hline
Tianshou & 8 & 5 & \cmark & \cmark & \cmark & 15-20\\ 
\hline
Coach & 11 & 8 & \cmark & \cmark & \xmark & N/A \\
\hline
ReAgent & 4 & 3 & \cmark & \cmark & \xmark & 5 \\
\hline
garage & 9 & 6 & \cmark & \cmark & \xmark & 5-10 \\
\hline
keras-rl & 3 & 5 & \cmark & \cmark & \cmark & 10-15\\
 \hline
MushroomRL & 9 & 7 & \cmark & \cmark & \xmark & 5-10 \\
\hline 
Tensorforce & 8 & 5 & \cmark & \cmark & \cmark & 5-15 \\
\hline 

\end{tabular}
\caption{Comparison of different DRL libraries.}
\label{table:algs&envs}
\end{table}

\textbf{Algorithms.} 
We first evaluate the algorithm support. All algorithms we considered in this comparison include DQN, HER, Rainbow~\cite{hessel2018rainbow}, vanilla policy gradient (VPG)~\cite{williams1992simple}, A2C, A3C, actor-critic with experience replay (ACER)~\cite{wang2016sample}, actor critic using Kronecker-factored trust region (ACKTR)~\cite{wu2017scalable}, DDPG, TD3, SAC, PPO, DPPO, TRPO, as well as the variants of DQN like double DQN~\cite{van2016deep}, dueling DQN~\cite{wang2015dueling}, Retrace~\cite{munos2016safe}, noisy DQN~\cite{fortunato2017noisy}, distributed DQN~\cite{bellemare2017distributional}, prioritized experience replay (PER)~\cite{schaul2015prioritized}, quantile regression DQN (QR-DQN)~\cite{dabney2018distributional}, $N$-step Q-learning~\cite{mnih2016asynchronous}, normalized advantage functions (NAF)~\cite{gu2016continuous} and Rainbow~\cite{hessel2018rainbow}.
As we can see from Table~\ref{table:algs&envs}, \sys supports
12 DRL algorithms, whereas Coach supports 11 algorithms and other libraries support less than 10 algorithms.
A key difference between \sys and other libraries is its support of
\emph{distributed} DRL algorithms, 
which makes \sys one of the few libraries that support distributed DRL algorithms
such as DPPO.
This type of algorithms is increasingly critical
because practitioners have recently achieved great success 
of training DRL agents using parallel learning framework~\cite{heess2017emergence}.

\vspace{0.1cm}
\noindent
\textbf{Environments.}
We then evaluate the environment support. The environments include: (1) Atari, Box2d, Classic control, MuJoCo, Robotics in OpenAI Gym (counted separately); (2) DeepMind Control Suite; (3) RLBench~\cite{james2020rlbench}; (4) Roboschool; (5) PyBullet~\cite{coumans2019}. As shown in Table~\ref{table:algs&envs},
\sys supports 7 environments, making it among
those libraries, e.g., Coach and MushroomRL, that provide a large collection of environments.
A key feature for \sys is its support for all observations types (e.g.,
Vector, Image, and Dictionary). The Dictionary in this paper indicates either a \emph{dictionary} or a \emph{tuple} type in Python, which is literally a collection of sub-data with different shapes. The other library: keras-rl, which can offer the same full support, only provide 3 DRL algorithms, whereas \sys can support 12 DRL algorithms. This shows the importance
of achieving automatic agent construction in \sys: new observations can be automatically supported by all DRL algorithms.
In addition,
the full observation support also makes \sys the only library, as far as we know,
that supports an important environment: RLbench.
This environment has growing popularity
due to the recent booming of robot learning applications.
It produces complex observations that contain
images, vectors, and dictionaries, making it difficult to be supported by existing libraries.

\vspace{0.1cm}
\noindent
\textbf{API expressiveness.}
We evaluate the API design by counting
the lines of code (LoC) for declaring DRL agents.
We exclude Baselines and Coach because they have only command-line interfaces. 
The LoCs here only consider necessary code for declaring agents, excluding other lines for importing libraries or assigning values for variables.
As we can see in Table~\ref{table:algs&envs}, \sys requires
4 LoCs to declare DRL agent while the ReAgent library comes as the second, costing 5 LoCs on average. Other programmable DRL libraries require
users to write around 10 - 20 LoCs. 
In addition, \sys differentiates with other libraries
in terms of its support for customising
agents. This makes \sys an attractive option for robot learning users who
often need to (i)~deal with RGB-D camera produced by the learning environment RLBench,
and (ii)~adopt customised network architectures like recurrent layers.




A complete comparison of RLzoo with other popular DRL libraries on (1) supported RL algorithms, (2) supported environments and (3) LoC are provided in Appendix~\ref{app:compare_table}.

\section{Conclusion}
This paper introduces \sys, a novel DRL library that makes
the development of DRL agents efficient.
\sys provides high-level yet flexible
APIs for declaring DRL agents. These APIs are particularly efficient 
in prototyping DRL agents, and scaling out the training of agents
to many nodes.
\sys further comes with a model zoo, enabling developers
to easily evaluate different DRL algorithms. 
In the future, we will consistently improve the API design of \sys, e.g.,
providing better support for implementing emerging multi-agent DRL algorithms. We will also add new DRL algorithms into the model zoo, especially
those targeting robot learning.

\section{Acknowledgements}
This project was supported by National Key R\&D Program of China: New Generation Artificial Intelligence Open Source Community and Evaluation (No.2020AAA0103500), Topic: New Generation Artificial Intelligence Open Source Community Software and Hardware Infrastructure Support Platform (No.2020AAA0103501). 

\bibliographystyle{unsrt}
\bibliography{reference}

\newpage
\begin{appendices}
\section{Comparison Table}
\label{app:compare_table}

\begin{table*}[htbp]
\begin{tabular}{ |p{1.9cm}||p{0.8cm}|p{1.2cm}|p{1.1cm}|p{0.9cm}|p{1.2cm}|p{0.9cm}|p{1.1cm}|p{1.8cm}| p{1.4cm}| }
 \hline
  & RLzoo & Baselines & Tianshou & Coach & ReAgent & garage & keras-rl & MushroomRL & Tensorforce\\
 \hline
 DQN   &  \cmark    & \cmark &   \cmark  &\cmark &\cmark &\cmark &\cmark &\cmark&\cmark\\
 DQN variants  &  4 & 4 &   2  & 8 & 4 & 1 & 3 & 4 & 4\\
 HER  &  \cmark  & \cmark & \xmark  &\cmark &\xmark &\cmark &\xmark&\xmark&\xmark\\
 Rainbow & \xmark  & \xmark&   \xmark &\cmark &\xmark &\xmark &\xmark&\xmark&\xmark\\
 VPG   & \cmark   & \xmark&   \cmark &\cmark &\xmark &\cmark &\xmark&\cmark&\cmark\\
 A2C &\cmark  & \cmark &   \cmark &\cmark &\xmark &\xmark &\xmark&\cmark&\cmark\\
 A3C &\cmark  & \xmark &   \xmark &\cmark &\xmark &\xmark &\xmark&\xmark&\cmark\\
 ACER  &  \xmark    & \cmark & \xmark   &\cmark &\xmark &\xmark &\xmark&\xmark&\xmark\\
 ACKTR  &  \xmark    & \cmark & \xmark  &\xmark &\xmark &\xmark &\xmark&\xmark&\xmark\\
 DDPG &  \cmark  & \cmark &   \cmark &\cmark &\xmark &\cmark &\cmark&\cmark&\cmark\\
 TD3 & \cmark   & \xmark &   \cmark &\cmark &\cmark &\cmark&\xmark&\cmark&\xmark\\
 SAC & \cmark   & \xmark&   \cmark &\cmark &\cmark &\cmark&\xmark&\cmark&\xmark\\
 PPO & \cmark  & \cmark&   \cmark &\cmark &\xmark &\cmark&\xmark&\cmark&\cmark\\
 DPPO & \cmark  & \xmark&   \xmark &\xmark &\xmark &\xmark&\xmark&\xmark&\xmark\\
 TRPO & \cmark  & \cmark&   \xmark &\xmark &\xmark &\cmark&\xmark&\cmark&\cmark\\
 \hline
 \# of Alg. & 12  &   9 &   8 & 11 & 4 & 9 & 3 &9 & 8\\
 \hline
 
  Atari           &  \cmark  & \cmark &  \cmark &\cmark &\cmark &  \cmark &  \cmark & \cmark & \cmark\\
 Box2D           &  \cmark  & \cmark & \cmark  &\cmark &\cmark &  \cmark &  \cmark & \cmark & \cmark\\
 Classic &  \cmark  & \cmark & \cmark  &\cmark &\cmark &  \cmark &  \cmark & \cmark & \cmark\\
 MuJoCo          & \cmark   & \cmark & \cmark  &\cmark &\xmark &  \cmark &  \cmark &  \cmark & \cmark\\
 Robotics        & \cmark   & \cmark &  \xmark &\cmark &\xmark &  \cmark &  \cmark & \cmark & \cmark\\
 Control Suite   & \cmark   & \xmark &  \xmark &\cmark &\xmark & \cmark  & \xmark & \cmark & \xmark\\
 RLbench         & \cmark   & \xmark &  \xmark &\xmark &\xmark &\xmark   & \xmark & \xmark & \xmark\\
 Roboschool      & \xmark   & \xmark &  \xmark &\cmark &\xmark &\xmark   & \xmark & \xmark & \xmark\\
 PyBullet        & \xmark   & \xmark & \cmark  &\cmark &\xmark &\xmark   & \xmark & \cmark & \xmark\\
  \hline
 \# of Env.      & 7        &  5     &  5     &8      &3      &6        &5 & 7 & 5\\
 \hline
    LoC     & 4        &  N/A     &  15-20      & N/A      & 5     & 5-10        & 10-15 & 5-10 & 5-15\\
  \hline
\end{tabular}
\caption{A complete comparison of RLzoo with other popular DRL libraries on: (1) supported RL algorithms, (2) supported environments, and (3) LoC. In ``\# of Alg.'', all DQN variants are counted as one type.}
\label{table:all}
\end{table*}
\end{appendices}

\end{document}